\documentclass{gSCS2e}%

\usepackage{epstopdf}% To incorporate .eps illustrations using PDFLaTeX, etc.
\usepackage{subfigure}% Support for small, `sub' figures and tables

\theoremstyle{plain}
\newtheorem{theorem}{Theorem}[section]

\theoremstyle{definition}

\theoremstyle{remark}
\newtheorem{remark}[theorem]{Remark}

\usepackage{graphicx}
\usepackage{latexsym,amsthm,amsmath,amssymb,amsfonts,mathrsfs}

\usepackage{color}
\usepackage[utf8]{inputenc}
\usepackage{amssymb,todonotes}
\usepackage{multirow, paralist}
\usepackage{url,xspace}
\usepackage{algorithm}
\usepackage{algorithmicx} 
\usepackage{algpseudocode,todonotes}

\newcommand{\etal}{\emph{et\,al.\xspace}}
\newcommand{\set}[2]{\left\{#1\,\left|\,#2\right.\right\}}
\newcommand{\slhc}[1]{{\mathfrak{sl}\!\left(#1\right)}}
\newcommand{\slhcinv}[1]{{\mathfrak{sl}^{{-1}}\!\left(#1\right)}}
\newcommand{\card}[1]{\left|#1\right|}

\begin{document}

%\jvol{00} \jnum{00} \jyear{2015} \jmonth{January}

%\articletype{GUIDE}

\title{Maximum Likelihood Estimation for Single Linkage Hierarchical Clustering}

\author{
\name{Dekang Zhu\textsuperscript{a}$^{\ast}$\thanks{$^\ast$Corresponding author. Email: dekang.zhu@foxmail.com};
 Dan P. Guralnik\textsuperscript{b};  Xuezhi Wang\textsuperscript{c}; Xiang Li\textsuperscript{a}; Bill Moran\textsuperscript{c}}
\affil{\textsuperscript{a}School of Electronic Science \& Engineering, National University of Defense Technology, Changsha, China;
\textsuperscript{b}Electrical \& Systems Engineering, University of Pennsylvania, Philadelphia, United States;
\textsuperscript{c}Electrical \& Computer Engineering, RMIT University, Melbourne, Australia.
}
%\received{v4.0 released January 2015}
}

\maketitle
\begin{abstract} We derive a statistical model for estimation of a dendrogram from single linkage hierarchical clustering (SLHC) that takes account of uncertainty through noise or corruption in the measurements of separation of data. Our focus is on just the estimation of the hierarchy of partitions afforded by the dendrogram, rather than the heights in the latter. The concept of estimating this ``dendrogram structure'' is introduced, and an approximate maximum likelihood estimator (MLE) for the dendrogram structure is described. These ideas are illustrated by a simple Monte Carlo simulation that, at least for small data sets, suggests the method outperforms SLHC in the presence of noise.
\end{abstract}


\section{Introduction}
Distance-based clustering is the task of grouping objects by some measure of similarity, so that objects in the same group (or {\it cluster}) are more similar or closer (with respect to a prescribed notion of distance) than those in different clusters. Clustering is a common technique for statistical data analysis, widely used in data mining, machine learning, pattern recognition, image analysis, bioinformatics and cyber security. 

Conventional (``flat'', ``hard'') clustering methods accept a finite metric space $(O,d)$ as input and return a partition of $O$ as their output. Hierarchical clustering (HC) methods have a different philosophy: their output is an entire hierarchy of partitions, called a \emph{dendrogram}, capable of exhibiting multi-scale structure in the data set~\cite{carlsson2008persistent,Carlsson2010}. Rather than fixing  the required number of clusters in advance, as is common for many flat clustering algorithms, it is more informative to furnish a hierarchy of clusters, providing an opportunity to choose a partition at a scale most natural for the context of the task at hand.

Many HC methods require linkage functions to provide a measure of dissimilarity between clusters (see~\cite{Everitt-Cluster_Analysis,Mullner-modern_clustering_algorithms} for a fairly recent review). Some commonly used linkage functions are single linkage, complete linkage, average linkage, etc. The SLHC method, though suffering from the so called ``chaining effect'', remains popular for large scale applications~\cite{Jain:1999:DCR:331499.331504} because of the low complexity of implementing it using minimum spanning trees (MST)~\cite{Gower69}. This work relies on a particularly useful representation of dendrograms using ultra-metrics, introduced by Jardine and Sibson~\cite{Jardine71}. Their point of view enabled redefinition of HC methods as maps from the collection of finite metric spaces to the collection of finite ultra-metric spaces~\cite{carlsson2008persistent,Carlsson_Memoli-classifying_clustering_schemes}. This enables a discussion of two essential properties~ ---~ stability and convergence with respect to the Gromov-Hausdorff metric~ ---~ that characterize SLHC within a broad class of HC methods~\cite{Carlsson2010}.

\textbf{Motivation}: As described in \cite{Sturmfels-biology_and_new_theorems}, distance-based clustering methods, hierarchical as well as flat and overlapping, are deeply rooted in several mathematical disciplines, and are ubiquitous in bio-informatics applications. For example, in contemporary applications to the analysis of gene expression data~\cite{shannon2003analyzing,butte2002use,levenstien2003statistical}, the raw data generated by microarrays is usually preprocessed to extract normalized expression values from which distance measures are computed, to be subsequently fed as input to a clustering algorithm. Depending on the kind of information sought, different variants of the conventional HC methods are applied, such as, for instance, hybrid HC~\cite{chipman2006hybrid} or improved Pearson correlation proximity-based HC~\cite{booma2014improved}. 

More generally, HC methods play an important role wherever learning and analysis of data have to be performed in an unsupervised fashion. For example, clustering is a key underpinning technology in most algorithms for cyber-security. In this context, clustering arises in a large number of applications, including malware detection~\cite{Wang_Miller_Kesidis-cluster_hierarchy_for_anomaly_detection}, identification of compromised domains in DNS traffic~\cite{biggio2014poisoning}, classification of sources and methods of attacks~\cite{4028522}, identification of hacker and cyber-criminal communities~\cite{lu10:_social_networ_analy_crimin_hacker_commun}, detection of repackaged code in applications for mobile devices~\cite{Hanna:2012:JSS:2481803.2481809}, and classification of files and documents~\cite{Steinbach2000}. There is an urgent need for more robust and reliable clustering algorithms.

Essentially all approaches to clustering, hierarchical or otherwise, accept the distances as ``the truth''. It is assumed uncorrupted by noise or artifacts. Particularly at the level of analogue data such as timing and device dependent parameters, but also even with some digital data, this is far from a correct model. For example, measures of dissimilarity between code samples are often engineered to reflect an opinion of the algorithm designer regarding the significance of specific features of executable code; it is more plausible to treat distance measures produced in this way as (quantifiably) uncertain measurements of the code sample rather than regard them as an objective truth.

Thus, practical necessities lead us to require that the output of clustering algorithms should account for uncertainty in the distance data and, to do that, a rigorous statistical approach is required. Obtaining dendrograms through statistical estimation (with an appropriate noise model for the data) will, in principle, result in improved HC methods to meet the needs of applications.

Conventional approaches to statistical estimation of partitions and hierarchies view the objects to be clustered as random samples of certain distributions over a prescribed geometry (e.g. Gaussian mixture model estimation using expectation-maximization in Euclidean spaces), and clusters can then easily be described in terms of their most likely origin. Thus, these are really {\it distribution-based} clustering methods~ ---~ not distance-based ones. Our approach directly attributes uncertainty to the process of obtaining values for the pairwise distances rather than distort the data by mapping it into one's ``favorite space''. To the best of our knowledge, very little work has been done in this vein. Of note is~\cite{Castro04likelihoodbased}, where similar ideas have been applied to the estimation of spanning trees in a communication network~(see related work below).

\textbf{Related work}: In phylogenetic
applications, the use of MLE and Bayesian methods for the estimation
of evolutionary trees is a time-honored tradition spanning decades~\cite{Cavalli_Edwards67,edwards1970estimation,Felsenstein73discrete,Felsenstein_73,Felsenstein:1981wd,Yang:1994eu,Rannala:1996qe,Yang01071997}, and various clustering methods having been introduced for purposes of ``phenetic clustering''~\cite{farris1972estimating,felsenstein1984distance}. In a rough outline, one estimates a tree structure to describe a population of samples from distributions of the form $p(\bm{x}|\tau)$, where $\tau$ is the evolutionary tree structure and the measurement $\bm{x}$ is a gene character (such as gene or nucleotide frequency); alternatively, one assigns (deterministically!) distance measures to reflect uncertainties in the quantities measured in the population. Estimation relying {\it only} on the noise model of the underlying dissimilarity measure would clearly constitute a much more general apparatus, compressing all the uncertainty about the data into the noise model, but otherwise treating all data sources with equal mathematical rigor. 

A serious hurdle in the way of brute-force MLE estimation of dendrogram structure is the super-exponential growth of the number of such structures with the size of the data set. Naturally, this aspect of the estimation problem is more readily seen in applications related to cyber-space, where large data sets dominate the scene. The work of Castro~\etal~\cite{Castro:2003fk,Castro04likelihoodbased} needs to be credited for having inspired an MCMC-based hypothesis-pruning procedure we have applied in this paper. However, we point out that both the clustering and the estimation problems that are the foci of their work are quite different from ours, and much more limited in scope. First, and most important, is that Castro~\etal~restrict attention to similarities with constant inter-cluster values, which effectively corresponds to postulating an ultra-metric setting \emph{ab initio}. This is well-suited for the purposes of their application (network topology identification), but is unsatisfactory for the general case. Secondly, the network model of Castro~\etal~is not, strictly speaking a metric model, as they do not enforce the constraints coming from the triangle inequalities. The notorious  complexity \cite{Deza_et_al,Deza_Laurent-cutbook} of this set of inequalities poses significant additional challenges to the problem of estimating structure from a measurement of a metric.

\section{Preliminaries}
\noindent{\bf Distance-Based Clustering. } Given a set of objects $O$, a hard clustering method generates a partition of $O$~ ---~ a collection of pairwise-disjoint subsets (\emph{clusters}) of $O$ whose union is $O$. For any $x\in O$ and any partition $R$ of $O$, we denote the cluster of $R$ containing $x$ by $R_x$.

We focus on \emph{distance-based clustering methods}, where a data set $O$ undergoes initial processing to produce a symmetric, real-valued, non-negative function $d$ on $O\times O$, whose values $d_{xy}$ satisfy the triangle inequality, and serve to quantify the ``degree of dissimilarity'' between data entries. In applications, the user has some freedom to determine the values of the $d_{xy}$ according to the requirements of the application in hand.

\smallskip
\noindent{\bf Hierarchical Clustering. } Attempts have been made to anchor distance-based clustering in a firm axiomatic foundation, but results so far have been negative: \cite{Kleinberg-impossibility} studies a seemingly intuitive and minimalistic axiomatic system for distance-based clustering which fails to support any clustering method; \cite{Meilua-comparing_clusterings_axiomatic} works in a similar vein to demonstrate that reasonable axiomatic notions of distance between outputs of a flat clustering method are equally elusive. As later described in \cite{carlsson2008persistent}, the key obstruction to such axiomatic approaches lies with the requirement to produce a single partition as its output. To resolve this issue they proposed HC methods, producing \emph{dendrograms}.

\begin{figure}[ht]
	\centering
		\includegraphics[width=.5\columnwidth]{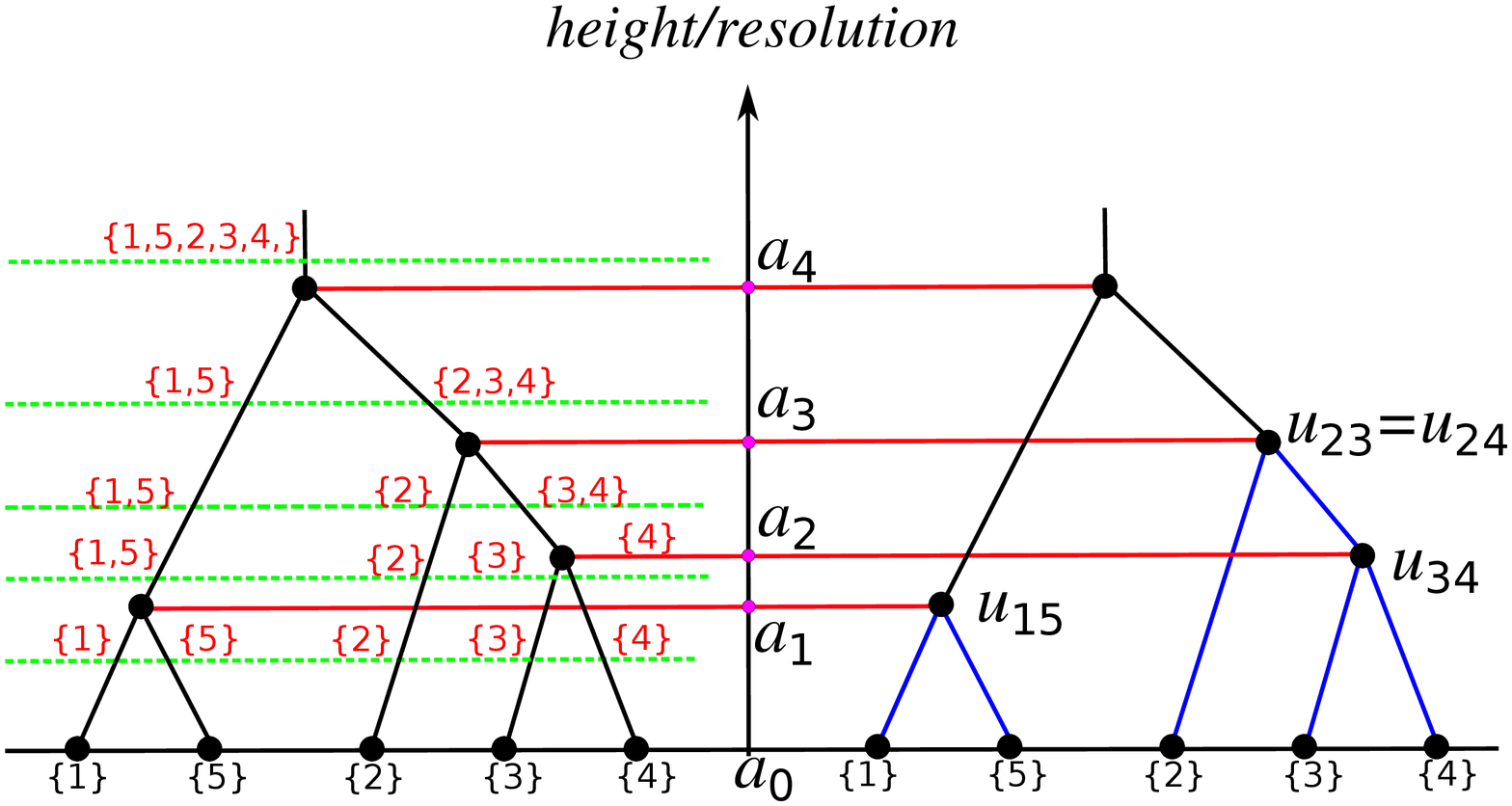}
		\caption{A rooted tree with labelled leaves as a dendrogram (left) and as an ultra-metric on $O=\{1,\ldots,5\}$ (right)}\label{fig:dendrogram versions}
\end{figure}

Recall that a partition $R_1$ is said to \emph{refine} a partition $R_2$, if every cluster of $R_1$ is contained in a cluster of $R_2$; we denote this by $R_1\succeq R_2$. A \emph{hierarchy} is a collection $\mathcal{C}$ of partitions, where every $R_1,R_2\in\mathcal{C}$ satisfy either $R_1\succeq R_2$ or $R_2\succeq R_1$. Intuitively, a dendrogram is a hierarchy with assigned heights, or \emph{resolutions}; this is usually represented visually as a rooted tree~ ---~ see Figure~\ref{fig:dendrogram versions} (left).

Formally, following \cite{Carlsson2010}, we describe a dendrogram as a pair $(O,\beta)$, where $\beta$ is a map of $[0,\infty)$ to the collection of partitions on $O$ satisfying the following:
\begin{compactitem}
	\item There exists $r_0$ such that $\beta(r)=\{O\}$ for all $r\geqslant r_0$;
	\item If $r_1\leqslant r_2$ then $\beta(r_1)$ refines $\beta(r_2)$;
	\item For all $r$, there exists $\epsilon>0$ s.t. $\beta(t) = \beta(r)$ for $t\in[r,r+\epsilon]$.
\end{compactitem}
Clusters of $\beta(r)$ are called  \emph{clusters at resolution $r$}.

\smallskip
\noindent{\bf Encoding Dendrograms. } Ultra-metrics provide a
convenient tool for encoding dendrograms~\cite{Jardine71}. Recall that
a metric $d$ on $O$ is said to be an \emph{ultra-metric}, if 
\begin{equation}
d_{xy} \leqslant \text{max}(d_{xz},d_{zy})\,,\quad \forall\; x,y,z\in O.
\end{equation}
The correspondence between dendrograms and ultra-metrics~\cite{Carlsson2010} is described as follows: any dendrogram $\beta$ gives rise to an ultra-metric $u=u(\beta)$, as shown in Figure~\ref{fig:dendrogram versions}: \vspace{-.02in}
\begin{equation}
	u(\beta)_{xy}:=\inf\set{r>0}{\beta(r)_x=\beta(r)_y}.
\end{equation}
Conversely, the dendrogram $\beta$ may be reconstructed from an ultra-metric $u$ by setting
\begin{equation}
	\beta(r)_x:=\set{y\in O}{u_{xy}\leq r}\,.
\end{equation}

\smallskip
\noindent{\bf Single Linkage Hierarchical Clustering (SLHC). } Single-linkage hierarchical clustering is defined, from the point of view of dendrograms, as follows. Given the metric space $(O,d)$, for each $r\geq 0$, a dendrogram $\theta_d$ is constructed by setting $x,y\in O$ to lie in the same cluster of the partition $\theta_d(r)$ if and only if there exists a finite sequence of points $x_0,\ldots,x_m\in O$ with $x_0=x$, $x_m=y$ and $d(x_{i-1},x_i)\leq r$ for all $i\in\{0,\ldots,m\}$. Such a sequence is called an \emph{$r$-chain from $x$ to $y$ in $(O,d)$}.

SLHC is often implemented by constructing an MST in $(O,d)$. The
partition $\theta_d(r)$ is obtained from \emph{any} MST $T$ of $(O,d)$
by removing all edges of $T$ of length$>r$; the clusters of the
corresponding dendrogram $\theta_d(r)$ are the connected components of
the resulting \emph{forest}, and the corresponding ultra-metric
distance $u_{xy}$ ($x,y\in O$) then equals the length of the longest
edge of $T$ (with respect to the distance $d$) separating $x$ from $y$
in $T$. Henceforth, we will write $u=\slhc{d}$ to denote the single
linkage mapping of a metric $d$ to the ultra-metric encoding of the
corresponding dendrogram $u$. 

It is a central result of \cite{Carlsson2010}, that SLHC is the unique hierarchical clustering method enjoying certain naturality properties, in stark contrast with the flat clustering situation. For a detailed discussion of the map $\slhc{\cdot}$, we refer the reader to \cite{carlsson2008persistent,Carlsson2010}.

\smallskip
\noindent{\bf Notation. } Note that metrics and ultra-metrics are conveniently written in matrix form, after fixing an order on $O$. Thus, writing $O:=\{o_1,\ldots,o_n\}$, we will use $\bm{\theta}:=[d_{ij}]$, $d_{ij}:=d(o_i,o_j)$ to denote the metric $d$ in matrix form, and $\bm{u}=[u_{ij}]$ to denote the ultra-metric obtained from it by applying the map $\slhc{\cdot}$.

\section{Statistical Model}

It is useful to separate the metric information in a dendrogram (the
grading by resolution) from the combinatorial information it conveys:
a dendrogram/ultra-metric $\bm{u}$ is uniquely represented by a pair
of  parameters $(\tau,\bm{a})$, where:
\begin{itemize}
\item The parameter, $\tau$, denotes the \emph{structure of $u$}: the hierarchy defined by $u$ (with the resolutions forgotten), ordered by refinement.
\item The parameter, $\bm{a}$, is the \emph{height vector of $u$}, whose coordinates, in order, indicate the minimum resolution at which each partition in the structure occurs~ ---~ see Figure~\ref{fig:dendrogram versions}.
\end{itemize}

%\begin{figure}[!t]
%\centering
%\includegraphics[height=0.3\columnwidth,width=.6\columnwidth]{Fig2}
%\caption{Two dendrogram structures of a same \Dekang{binary tree}}
%\centering
%\label{LabeledHistories}
%\end{figure}

In what amounts to a choice of scale, we focus attention on the subset  $\Theta$ of the space of all metrics $\theta$ satisfying $\slhc{\bm{\theta}}_{ij}\leq 1$ for all $1\leq i,j\leq n$. This is a compact convex set in $\mathbb R^{n(n-1)/2}$.
Restricting attention to $\bm{\theta}\in\Theta$ is equivalent to placing a restriction on $\bm{a}$ to lie in the set  $\Omega$ of all vectors $\bm{a}$ satisfying $0 \leqslant a_1 \leqslant a_2 \leqslant \cdots \leqslant a_{n-1}\leqslant 1$. Note that $\Theta$ coincides with the pre-image under $\slhc{\cdot}$ of the set of all ultra-metrics $\bm{u}$ with $\bm{a}\in\Omega$.

\begin{remark} It must be observed that {\it degenerate structures};
  that is,   structures containing fewer than $n=\card{O}$ partitions (or,
  equivalently, corresponding to dendrograms that are not binary
  trees), occur in a set of metrics of Lebesgue measure zero, and therefore do not have any effect on statistical considerations regarding SLHC. Other clustering algorithms, such as hierarchical $2$-means~\cite{Arslan_Guralnik_Koditschek-HC_navigation1,Arslan_Guralnik_Koditschek-HC_navigation2}, for example, do not have this property and, therefore, require more delicate analysis.
\end{remark}

Our statistical model, introduced in~\cite{dekang2015_1}, is as follows:
\begin{itemize}
	\item The measurement $\bm{x}$ only depends on a metric $\bm{\theta}\in\Theta$ through a specific distribution $p(\bm{X}|\bm{\theta})$. 
	\item The ultra-metric $\bm{u}=\slhc{\bm{\theta}}$ is the parameter to be estimated from $\bm{x}$, with unknown $\bm{\theta}$ playing the rule of nuisance (latent) parameter. 
	\item A reasonable assumption for this noise model is that the measurements of the different values of $\theta$ are sampled independently from the same parametrized distribution $p(X|\theta)=G_\theta(X)$, $\theta\in(0,+\infty)$:
	\begin{equation}
		p(\bm{X}|\bm{\theta}) = \prod _{ij} G_{\theta_{ij}}(X_{ij})\,;
	\end{equation}
\end{itemize}
Following the recommendations of~\cite{berger1999integrated}, we pick integrated likelihood for our method of eliminating the nuisance parameter $\bm{\theta}$. Given a measurement $\bm{x}$ the likelihood function is:
\begin{equation}\label{eq:inner_integral}
	\mathcal{L}(\bm{u};\bm{x}):=p(\bm{x}|\bm{u})
	=\int p(\bm{x}|\bm{\theta}) p(\bm{\theta}|\bm{u})\textbf{d}{\bm{\theta}}
	=\int \prod _{ij} G_{\theta_{ij}}(x_{ij}) p(\bm{\theta}|\bm{u})\textbf{d}{\bm{\theta}} 
\end{equation} 
In the context of our problem the support of $G_\theta$ is restricted to $(0,+\infty)$.

%Our statistical model is as follows: the data is provided by a measurement $\bm{x}$ of the distance matrix $\bm{\theta}\in\Theta$, with the ultra-metric $\bm{u}$ playing the role of the parameter to be estimated. The measurement $\bm{x}$ only depends on $\bm{\theta}$ through a specific statistical distribution; the metric $\bm{\theta}$ is dependent on the ultra-metric $\bm{u}$. The likelihood for the estimation problem is
%\begin{equation}\label{eq:1} 
%	p(\bm{x},\bm{\theta}|\bm{u})=p(\bm{x}|\bm{\theta})p(\bm{\theta}|\bm{u})\,, 
%\end{equation}
%where $\bm{\theta}$ is a nuisance parameter. Marginalizing with respect to $\theta$ gives
%\begin{equation}\label{eq:inner_integral}
%	p(\bm{x}|\bm{u})  =  \int p(\bm{x}|\bm{\theta}) p(\bm{\theta}|\bm{u}) \textbf{d}\theta.
%\end{equation}

The height vector $\bm{a}$ is implicit in the likelihood  function, because of the complex integral, and so we focus on estimating the dendrogram structure $\tau$, treating height vector $\bm{a}$ as a nuisance parameter for this task. It is reasonable to assume that the structure and height functions are independent parameters. Replacing $\bm{u}$ by $(\tau,\bm{a})$, we obtain
\begin{equation}\label{eq:closed_form_expressionx_pxtau}
	p(\bm{x}|\tau) = \int p(\bm{x}|\bm{u})p(\bm{a}) \textbf{d}a = \int p(\bm{x}|\tau,\bm{a}) p(\bm{a}) \textbf{d}a .
\end{equation}
The MLE dendrogram structure is given by
\begin{equation}
\hat{\tau}(\bm{x}) = \text{arg } \underset{\tau}{\text{max }} p(\bm{x}|\tau).
\end{equation}

\subsection{The likelihood of the metric given the data} Since entries of $\bm{x}$ are measurements of distance, they are assumed to satisfy $x_{ij}>0$ and, for the purposes of this paper, we assume each $x_{ij}$ follows a log-normal distribution $\text{ln}\mathcal{N}(\mu_{ij},\sigma_{ij})$. Other noise distributions could be considered here. Therefore, 
\begin{equation}\label{eq:data_model_probability}
	p(\bm{x}|\bm{\theta})=
	\prod _{1\leq i<j \leq n}
		\frac{1}{\sigma_{ij} x_{ij} \sqrt{2\pi}}
		\text{exp}\left(
			-\frac{(\text{ln}x_{ij} - \mu_{ij})^2}{2\sigma_{ij} ^2}
		\right),
\end{equation}
where $\mu_{ij}$ and $\sigma_{ij}$ are distribution parameters. We model the relationship between the measurement and the true metric $\bm{\theta}$ as follows: we require $\text{E}[x_{ij}]=\theta_{ij}$ and $\text{Var}[x_{ij}]= v$, which sets the distribution parameters accordingly to be
\begin{equation}
	\mu_{ij} =  \text{ln}\theta_{ij} - \frac{1}{2}\text{ln}(1+\frac{v}{\theta_{ij}^2})\,,\quad
	\sigma_{ij} = \sqrt{\text{ln} (1 + \frac{v}{\theta_{ij}^2})}\,.
\end{equation}

\subsection{The likelihood of the dendrogram given the metric} In the absence of assumptions on the latent parameters $\bm{\theta}$, $p(\bm{\theta}|\bm{u})$ is taken to be a uniform distribution on its support. This is justified by the fact that, considering the total weight/length of an MST as a natural energy functional on the space of metrics, it follows directly from \eqref{eq:division_polytope} below that a maximum entropy distribution (subject to a total energy constraint) on this space restricts to a uniform distribution on the pre-image $\slhcinv{\bm{u}}$ of $\bm{u}$ under the single linkage map. Hence:
\begin{equation}
	p(\bm{\theta}|\bm{u}) = \left\{\begin{array}{cl}
		\frac{1}{\text{Vol}\left (\slhcinv{\bm{u}}\right)}
			&\text{ when } \bm{\theta} \in \slhcinv{\bm{u}}\\
		0	&\text{ otherwise }
	\end{array}\right. 
\end{equation}								
In more detail, following \cite{dekang2015_1}, denote 
\begin{equation}
	\text{MST}(\bm{\theta})=\set{T}{T \text{  is an MST of }\bm{\theta}}.
\end{equation}
For any spanning tree of $K_O$, the complete graph with vertex set $O$, write 
\begin{equation}
	C(T):=\set{\bm{\theta}}{\bm{\theta} \text{ is a metric on $O$ with }T\in\text{MST}(\bm{\theta})}
\end{equation}
and observe that any two of the cones $C(T)$ intersect in a set of measure zero. In addition, the following identity is proved in \cite{dekang2015_1}:
\begin{equation}\label{eq:division_polytope}
	\slhcinv{\bm{u}} = \bigcup_{T \in \text{MST}(\bm{u})}  C(T,\bm{u})
\end{equation}
where $C(T,\bm{u})$ is defined to be the set of all metrics $\bm{\theta}$ in $C(T)$ which coincide with $\bm{u}$ on the edges of $T$. In particular, any integration over $\slhcinv{\bm{u}}$ decomposes as a sum of integrals over the relevant $C(T,\bm{u})$. It is easy to verify that each $C(T,\bm{u})$ is a polytope given by the inequalities
\begin{equation}\label{eq:inequalities_for_bound}
	C(T,\bm{u}):\left\{ \begin{array}{l}
		\theta_{ij}-\theta_{ik}-\theta_{kj} \leqslant 0\\ 
		-\theta_{ij} \leqslant -u_{ij}\\
		\theta_{ij} \leqslant U_{ij}
	\end{array} \right.
\end{equation}
where $U_{ij}$ is defined to be the sum of the $\bm{u}$-lengths of all
the edges separating $o_i$ from $o_j$ in the tree $T$ for all
$i,j,k$. Since membership  in $C(T,\bm{u})$ is easily verified
using the inequalities in \eqref{eq:inequalities_for_bound}, this
enables   Monte-Carlo evaluation of integrals over these domains.

\subsection{The Prior on the Space of Dendrograms} We will assume that the parameter $\bm{a}$ is uniformly distributed in $\Omega$. As a result,
\begin{equation}
	p(\bm{a}) = \left\{\begin{array}{cl}
		\displaystyle\frac{1}{\text{Vol}(\Omega)}	&\text{ when } \bm{a} \in \Omega\\
		0 	&\text{otherwise}
	\end{array}\right.
\end{equation}

Thus, computing $p(\bm{x}|\tau)$ using
Equation~\eqref{eq:closed_form_expressionx_pxtau} requires the
computation of integrals both over $\Omega$ and over the polytopes
$C(T,\bm{u})$. Existing techniques for these computations, especially
the latter, are computationally extremely complex~\cite{barvinok1993computing,barvinok2010maximum,lasserre01laplace}. In this paper we resort to numerical approximation techniques to be introduced in the next section.

\section{Monte Carlo integration}
Monte Carlo integration~\cite{mackay1998introduction} is applied here to approximate the integrals.

\subsection{Equation~\eqref{eq:inner_integral}} \noindent{\bf Setting up the Approximation. } The inner integral $p(\bm{x}|\bm{u})$ splits, by \eqref{eq:division_polytope}, as a sum of integrals computed separately over each polytope $C(T,\bm{u})$. More precisely, we enumerate the MSTs associated with $\bm{u}$, $\text{MST}(\bm{u})=\{T_k\}_{k=1}^{K}$ where $K$ may depend on $\bm{u}$, and set $C_k=C(T_k,\bm{u})$. Then $p(\bm{x}|\bm{u})$ becomes
\begin{equation}\label{eq:numericalFormInteg}
	p(\bm{x}|\bm{u}) = \int\limits_{\slhcinv{\bm{u}}}\!\!\!\!\!\!\!
		p(\bm{x}|\bm{\theta}) p(\bm{\theta}|\bm{u}) \textbf{d}\theta =
	\sum_{k = 1}^{K} p(T_{k}|\bm{u})\!\!
		\int\limits_{C_k}\!
			p(\bm{x}|\bm{\theta})p(\bm{\theta}|T_k)\textbf{d}\theta .
\end{equation}
Since $\bm{\theta}$ is uniformly distributed in $\slhcinv{\bm{u}}$, $p(\bm{\theta}|T_k)$ is uniform in $C_k$ and $p(T_k|\bm{u})$ equals $\frac{\text{Vol}(C_k)}{\text{Vol}(\slhcinv{\bm{u}})}$.

For the Monte-Carlo approximation we  draw $N$ samples
$\{\bm{\theta}^{(l)}\}_{l=1}^N$ of metrics uniformly  from $\slhcinv{\bm{u}}$, with $N_k$ of these metrics drawn from each $C_k$. Then the Monte-Carlo approximation  of $p(T_k|\bm{u})$ is $N_k/N$ and, since $\sum_{k}N_k=N$, the Monte Carlo integration carried out for \eqref{eq:numericalFormInteg} yields the expression:
\begin{equation}\label{eq:numericalFormFinal}
	\phi(\bm{a}):=\sum_{k=1}^{K}\frac{N_k}{N}\cdot\frac{1}{N_k}\sum_{\bm{\theta}^{(l)}\in C_k}p\left (\bm{x}|\bm{\theta}^{(l)}\right) = \frac{1}{N} \sum_{l=1}^{N}p\left( \bm{x}|\bm{\theta}^{(l)} \right)  .
\end{equation}

\noindent{\bf Drawing the Samples. } According to Equation~\eqref{eq:inequalities_for_bound}, lower and upper bounds on the value $\theta_{ij}$ of a metric $\bm{\theta} \in \slhcinv{\bm{u}}$ are, respectively,
\begin{equation}
	\theta_{\text{min}}=\text{min}(\{u_{ij}\})\,,\quad
	\theta_{\text{max}}=\text{max}(\{U_{ij}\}) .
\end{equation}
Uniformly sampling metrics from the box defined by these bounds, and keeping only those lying in $\slhcinv{\bm{u}}$ as our samples for \eqref{eq:numericalFormFinal} is a straightforward approach, but  fails to produce enough samples, because $\slhcinv{\bm{u}}$ has measure zero in the space of all nonnegative symmetric dissimilarity matrices in $\mathbb R^{n(n-1)/2}$. Instead, we draw metrics separately in each $C_k$, $k\in\{1,\ldots,K\}$ as described below and then combine these counts:
\begin{itemize}
	\item{\bf Step 1. } Set $\theta_{ij} = u_{ij}$ when edge $e_{ij} \in T_k$, then uniformly draw values between $\theta_{\text{min}}$ and $\theta_{\text{max}}$ for the other parameters;
	\item{\bf Step 2. } Check the first constraint in \eqref{eq:inequalities_for_bound} (triangle inequality) on each drawn vector, keep a certain number of these  metrics;
	\item{\bf Step 3. } Check the last two constraints in \eqref{eq:inequalities_for_bound} for $T=T_k$, and keep the metrics satisfying them.
\end{itemize}

\subsection{Equation~\eqref{eq:closed_form_expressionx_pxtau}} 
\noindent{\bf Setting up the Approximation. } The Monte Carlo approximation for the likelihood function $\mathcal{L}(\tau;\bm{x})=p(\bm{x}|\tau)$ obtained from $N_\Omega$ samples $\{\bm{a}^{(l)}\}_{l=1}^{N_\Omega}$ drawn uniformly from $\Omega$ is:
\begin{equation}\label{eq:likelihood_outer_integral}
	\mathcal{L}(\tau;\bm{x}) \approx \int _{\bm{\Omega}} \phi(\bm{a}) p(\bm{a})\textbf{d}a \approx \frac{1}{N_\Omega}\sum _{l = 1}^{N_\Omega} \phi(\bm{a}^{(l)}) .
\end{equation}

\noindent{\bf Drawing the Samples.} Uniformly drawing a vector from
$\Omega$ is not so  straightforward a task. We observe that the
uniform distribution on the standard simplex $\Delta$ is a Dirichlet distribution~\cite{Ng_Tian_Tang-dirichlet_distributions} with parameter vector $(1,1,\ldots,1)$. 
This can be mapped to $\Omega$ using the linear change of variables replacing the standard basis $(\bm{e}_i)_{k=1}^n$ of $\mathbb{R}^n$ by $\left\{\bm{w}_k:=\sum_{i>k}\bm{e}_i\right\}_{k=1}^n$. Rewriting the vectors $\bm{a}\in\Omega$ in this basis yields:
\begin{equation}
	\bm{a} = \sum _{i = 1}^{n} \gamma _i w_i\,,\qquad
	\sum _{i = 1}^{n} \gamma _i = 1, \gamma _i \in [0,1] .
\end{equation}
Equivalently, for any $\bm{a}=(a_k)\in\Omega$, we can write it as $a_{k} = \sum _{i=1}^{k} \gamma _i$ for $\bm{\gamma}=(\gamma_k) \in \Delta$. This transformation is volume preserving, so that uniform sampling of $\bm{a}\in\Omega$ is equivalent to uniform sampling of $\bm{\gamma}\in\Delta$. This method produce samples of the required kind.

Algorithm~\ref{alg:numericallog-likelihood} gives the pseudo-code for the numerical integration.
\begin{algorithm}
\caption{Pseudo-code for approximate computing of the likelihood \eqref{eq:likelihood_outer_integral}}
\label{alg:numericallog-likelihood}
 \begin{algorithmic}[1]
 \Function{Compute}{$\bm{x}$, $\tau$};
  \For{$h = 1 : N_{\Omega}$};
  \State Draw a height vector $\bm{a}_h$ uniformly;
  \State $\bm{u}_{h} = (\tau,\bm{a}_h)$;
  \State Draw $\{ \bm{\theta}^{(l)} \} \in \slhcinv{\bm{u}_{h}}$ uniformly;
  \State Likelihood $prob(h) = \underset{l}{\text{mean }}p(\bm{x}|\bm{\theta}^{(l)})$;
  \EndFor
  \State \Return{log($\frac{\sum_h prob(h)}{N_\Omega}$)} .
 \EndFunction 
  \end{algorithmic} 
\end{algorithm}

\section{Reducing the Complexity}

There are $n!(n-1)!/2^{n-1}$ elements in the set of combinatorial
types of dendrograms on $n$ particles~\cite{edwards1970estimation}.
Denote the space of all such types by $\Lambda_n$. The explosive
growth of this set as a function of $n$ makes brute-force maximization
over it a prohibitive task even for reasonably small values of
$n$. Nevertheless, our data (see
Figures~\ref{fig:Appearance_ratio_0dian1_onefig}--\ref{fig:Rank_std_0dian3})
indicates  that structures of sufficiently low likelihood very rarely
coincide with the target structure. The removal from consideration of
such structures will result in little information loss for the outcome
of MLE while significantly reducing computational cost. This suggests
adaptation of  a similar approach to that
of~\cite{Castro04likelihoodbased} to produce an approximation of the
MLE estimator: for a fixed measurement $\bm{x}$, we regard the
likelihood $p(\bm{x}|\bm{\theta})$ as a distribution over $\Theta$ up
to a normalizing factor denoted by $\eta$ and draw a collection of
metrics $\{\bm{\theta}_k\}$ from $\Theta$ using the
Metropolis--Hastings (MH) algorithm~\cite{chib1995understanding} with
target distribution $\eta p(\bm{x}|\bm{\theta})$. From the resulting collection of structures, we choose the subset $\Lambda_s\subset \Lambda_n$ of those structures appearing with highest frequencies, and run the computation from the previous section only for structures $\tau\in\Lambda_s$. (intuitively, as long as $\bm{x}$ is a measurement of reasonable quality, more metrics in the set $\{\bm{\theta}_k\}$ are close enough to the true metric $\bm{\theta}$ in order for $\bm{\theta}_k$ to support the same structure as $\bm{\theta}$ does). 

For our implementation of the MH algorithm, we choose a proposal distribution
\begin{equation}\label{eq:p_thetanew_theta}
	g(\bm{\theta}^{\text{old}}\rightarrow\bm{\theta}^{\text{new}}) = \prod_{ij = 12}^{(n-1)n} \frac{1}{\sqrt{2\pi}\sigma} \text{exp}\left \{-\frac{(\theta_{ij}^{\text{\text{new}}} - \theta_{ij}^{\text{old}})^{2}}{2\sigma^2}\right \}
\end{equation}
with $\sigma = \sqrt{v}$,  keeping in mind the possibility that the sample $\bm{\theta}^{\text{new}}$ might not be a metric, in which case the sample will be discarded (thus, $g(\bm{\theta}^{\text{old}}\rightarrow\bm{\theta}^{\text{new}})$ is, in fact, zero in the complement of $\Theta$ and is otherwise only proportional to the above expression).

As the proposal distribution is symmetric, for the acceptance probability we may use the Metropolis Choice:
\begin{equation}\label{eqn:Metropolis choice}
	A(\bm{\theta}^{\text{old}}\rightarrow\bm{\theta}^{\text{new}}) = \text{min}\left(1,\frac{p(\bm{x}|\bm{\theta}^{\text{new}})}{p(\bm{x}|\bm{\theta}^{\text{old}})} \right)
\end{equation}

We recall that samples are generated iteratively. At each iteration, a new state is drawn from the proposal distribution for the current state. A real number $q$ is drawn uniformly at random from $[0,1]$, and the new state is accepted if $q\leq\frac{p(\bm{x}|\bm{\theta}^{\text{new}})}{p(\bm{x}|\bm{\theta}^{\text{old}})}$. Otherwise, the new state is rejected and the process remains in the same state. With additional burn-in and thinning, the iteration ends when a required number of metrics, $N_{\theta}$, is obtained. Algorithm \ref{alg:Pseudo-code_of_culling_process} summarizes the whole process. 

\begin{algorithm}
  \caption{Hypothesis pruning process for MLE estimation of dendrogram
    structure from a measurement $\bm{x}$ of a metric based on
    Metropolis-Hastings approximation of
    $\ell(\tau)\propto p(\bm{x}|\tau)$}
\label{alg:Pseudo-code_of_culling_process}
\begin{algorithmic}

	\Function{MH\_sampler}{$\bm{x}$}
		\State $\beta\gets$ duration of burn-in period
		\State $\delta k\gets$ thinning step
		\State $N_\theta\gets$ number of metrics to be sampled from $\Theta$
		\State $N_h\gets$ number of hypotheses for output
		\State $\theta_0\gets$ arbitrary element of $\Theta$
		
		\smallskip
		\For{$k=1\textbf{ to }\beta+\delta k\cdot N_\theta$}
			\State $\theta_{k}\gets$\Call{MH\_transition}{$\theta_{k-1}$}
		\EndFor
		
		\smallskip
		\State\Return{$N_h$ most frequently encountered structures\\ \hfill from $\set{\theta_k}{k-\beta\geq 0\,,\;\delta_k\text{ divides }(k-\beta)}$\phantom{$N_h mos$}}
		\smallskip	
	\EndFunction
	
	\medskip
	\Function{MH\_transition}{$\bm{\theta}^{\text{old}}$}
		\Repeat
			\State $\bm{\theta}^{\text{new}}\gets$ a draw from \eqref{eq:p_thetanew_theta}
		\Until{$\bm{\theta}^{\text{new}}\in\Theta$}
		\smallskip
		\State $A\gets\dfrac{p(\bm{x}|\bm{\theta}^{\text{new}})}{p(\bm{x}|\bm{\theta}^{\text{old}})}$
		\smallskip
		\State $u\gets$ draw from $\mathtt{Uniform(0,1)}$
		\smallskip
		\If{$u\leq A$}
			\State\Return{$\bm{\theta}^{\text{new}}$}
		\Else
			\State\Return{$\bm{\theta}^{\text{old}}$}
		\EndIf
	\EndFunction

\end{algorithmic} 
\end{algorithm}

\section{Simulation}
We demonstrate the effectiveness of the proposed hypothesis pruning process in the 5-particle case. A 5-point dendrogram may have any one of $5!\times 4!/2^{4}=180$ different dendrogram structures, which can be enumerated and indexed using the algorithm proposed in \cite{Yang01071997}.

First, we randomly draw a structure from $\Lambda_5$ and a height
vector from $\Omega$ to construct an objective ultra-metric (or,
equivalently, a dendrogram). Then a random metric $\bm{\theta}^\ast$
is sampled from the pre-image (under $\slhc{\cdot}$) of this
ultra-metric. This serves as the ground-truth metric used later to
generate one measurement with a specified noise level. Finally, we
implement MH as in Algorithm \ref{alg:Pseudo-code_of_culling_process}
with this measurement for its input to obtain a sequence of metrics
$\{\bm{\theta}_k\}$ from the distribution
$\eta p(\bm{x}|\bm{\theta})$. For each measurement $\bm{x}$ of
$\bm{\theta}^\ast$ $10000$ steps of Algorithm 2 are computed,
including $\beta=1000$ steps of 
burn-in, and applying a thinning of  $\delta_k = 3$ steps. The
resulting output of  $N_{\theta} = 3000$ observations is processed as
follows.  
\begin{enumerate}
	\item For each structure in $\Lambda_5$ its observed appearance frequency is calculated from among the $\{\bm{\theta}_k\}$ (the frequency is set to $0$ for structures which did not arise).
	\item We subdivide the range $[0,1]$ into 20 bins $[a_{i-1},a_i]$ of equal lengths, and generate a vector $\bm{v}$ of 20 integers, indicating for each bin $i$ the number of structures occurring with a frequency in $[a_{i-1},a_i]$.	
	\item This binning produces a ranking of the structures, in descending order, according to the frequency of occurrence.
\end{enumerate}

\begin{figure}%
\begin{center}
\parbox{0.45\columnwidth}{%
\includegraphics[width=0.4\columnwidth]{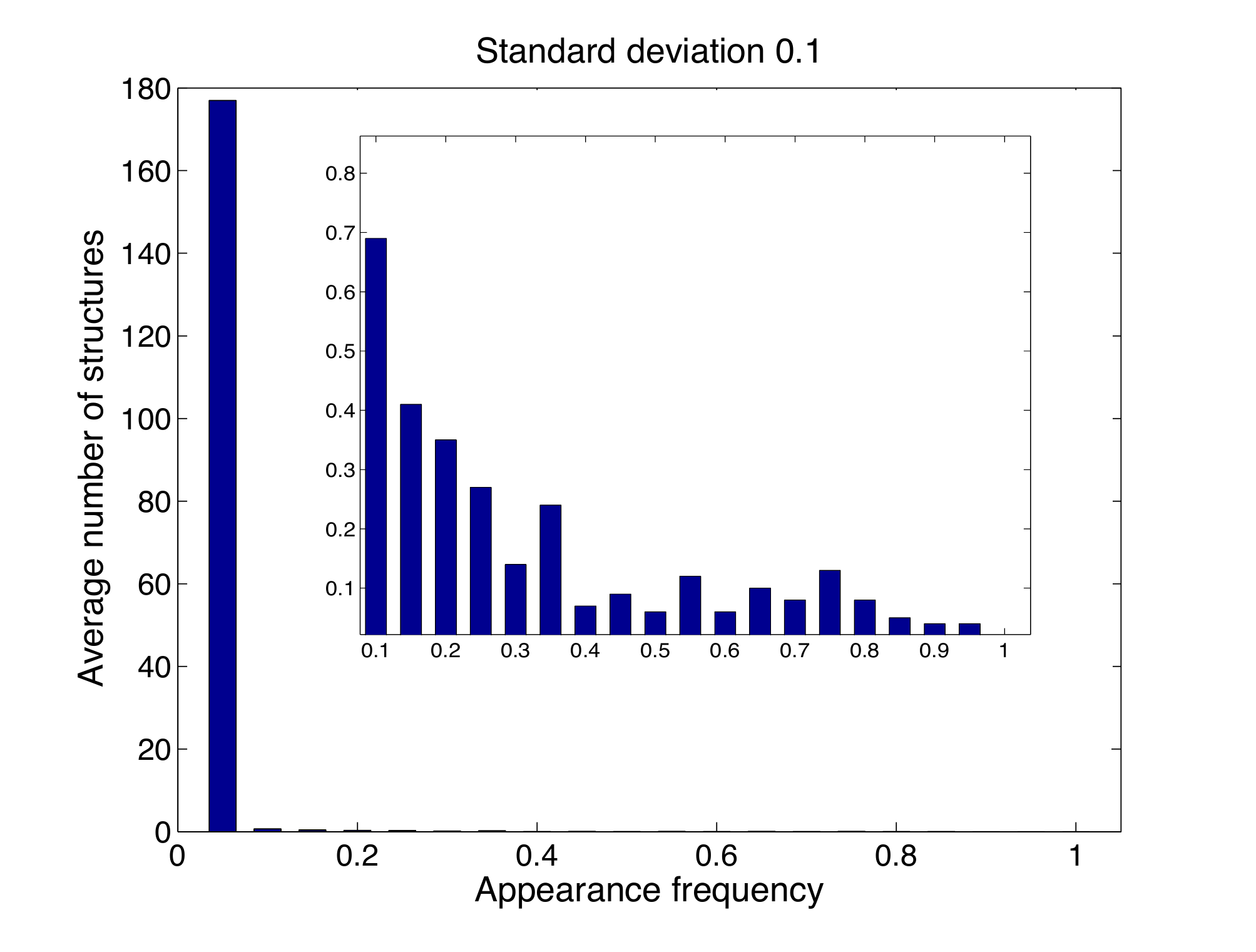}
\caption{Average number of structures in each interval with std 0.1.}%
\label{fig:Appearance_ratio_0dian1_onefig}}%
\qquad
\parbox{0.45\columnwidth}{%
\includegraphics[width=0.4\columnwidth]{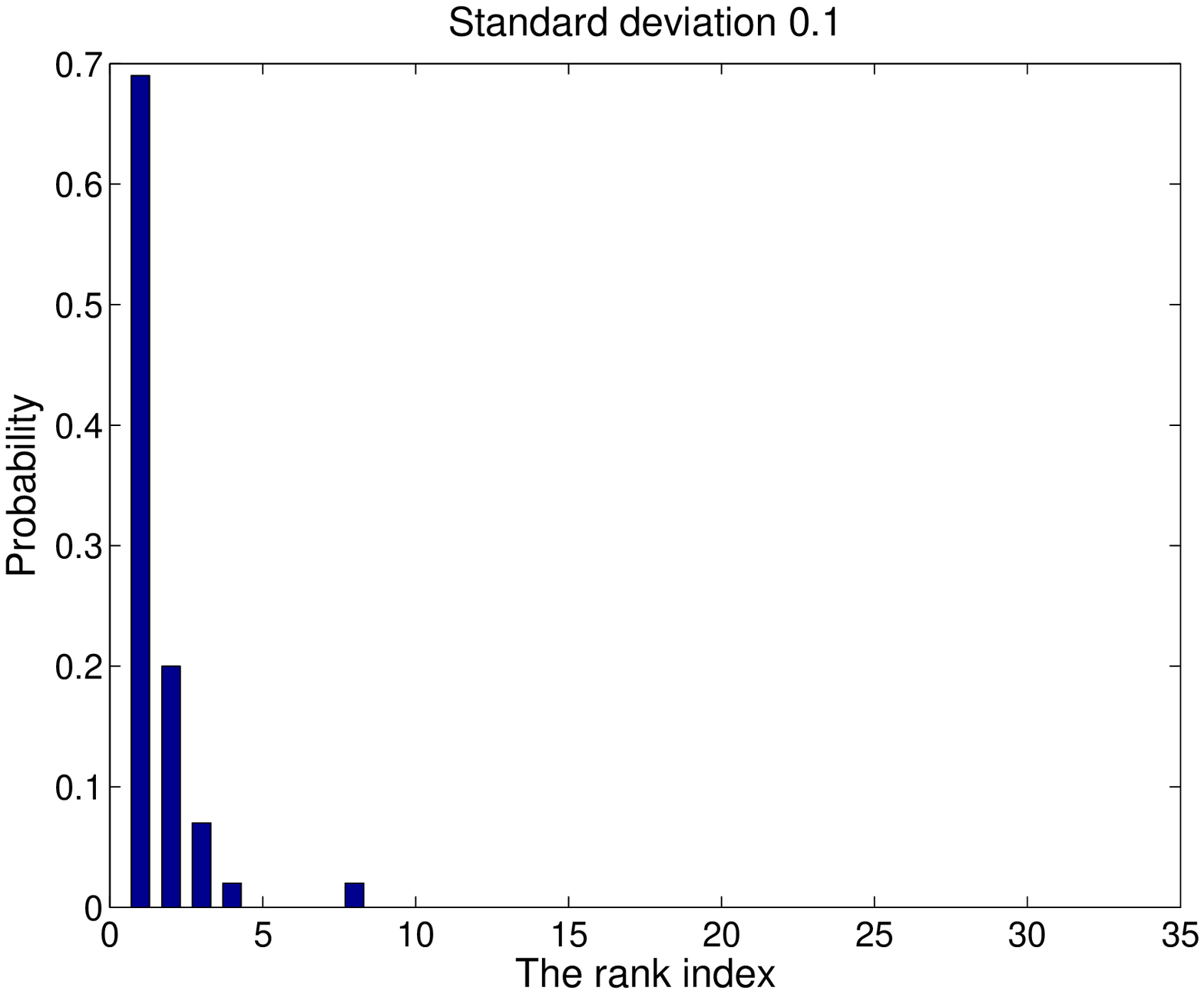}
\caption{The probability distribution of the objective structure's rank with std 0.1.}%
\centering
\label{fig:Rank_std_0dian1}}%
\end{center}
\end{figure}

Figure~\ref{fig:Appearance_ratio_0dian1_onefig} shows the averaged
histogram of the output vectors $\bm{v}$ generated from $1000$ iid
measurements $\bm{x}$ of $\bm{\theta}^\ast$ with standard
deviation $0.1$. Figure~\ref{fig:Rank_std_0dian1} shows the
corresponding distribution of the rank of the true
structure.

The inset in Figure~\ref{fig:Appearance_ratio_0dian1_onefig} provides an enlarged plot of the bars excluding the leftmost one: observe that more than 170 of the 180 possible structures have insignificant appearance ratios (in the interval $[0,0.05]$).

At the same time, Figure~\ref{fig:Rank_std_0dian1} shows that the rank of the true structure is almost completely distributed among the 10 highest ranked, with a nearly 70\% chance of the true structure ranking first. 

Figures~\ref{fig:Appearance_ratio_0dian1_onefig} and
~\ref{fig:Rank_std_0dian1} support our contention that most of the
structures may be removed from consideration, and that with little
chance of harm we may restrict attention to just a few of the highest ranking structures.

\begin{figure}%
\begin{center}
\parbox{0.45\columnwidth}{%
\includegraphics[width=0.4\columnwidth]{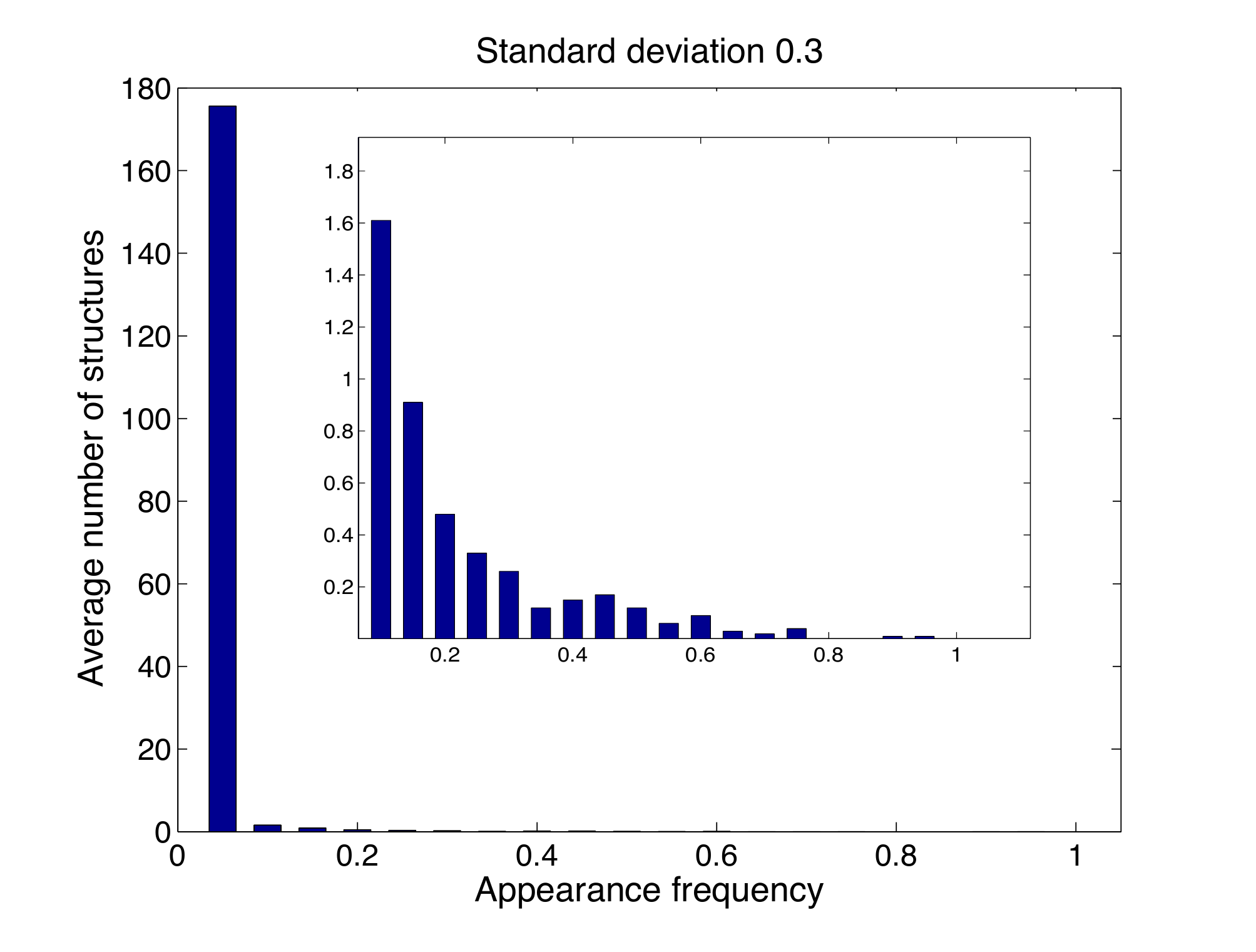}
\caption{Average number of structures in each interval with std 0.3.}%
\label{fig:Appearance_ratio_0dian3_onefig}}%
\qquad
\parbox{0.45\columnwidth}{%
\includegraphics[width=0.4\columnwidth]{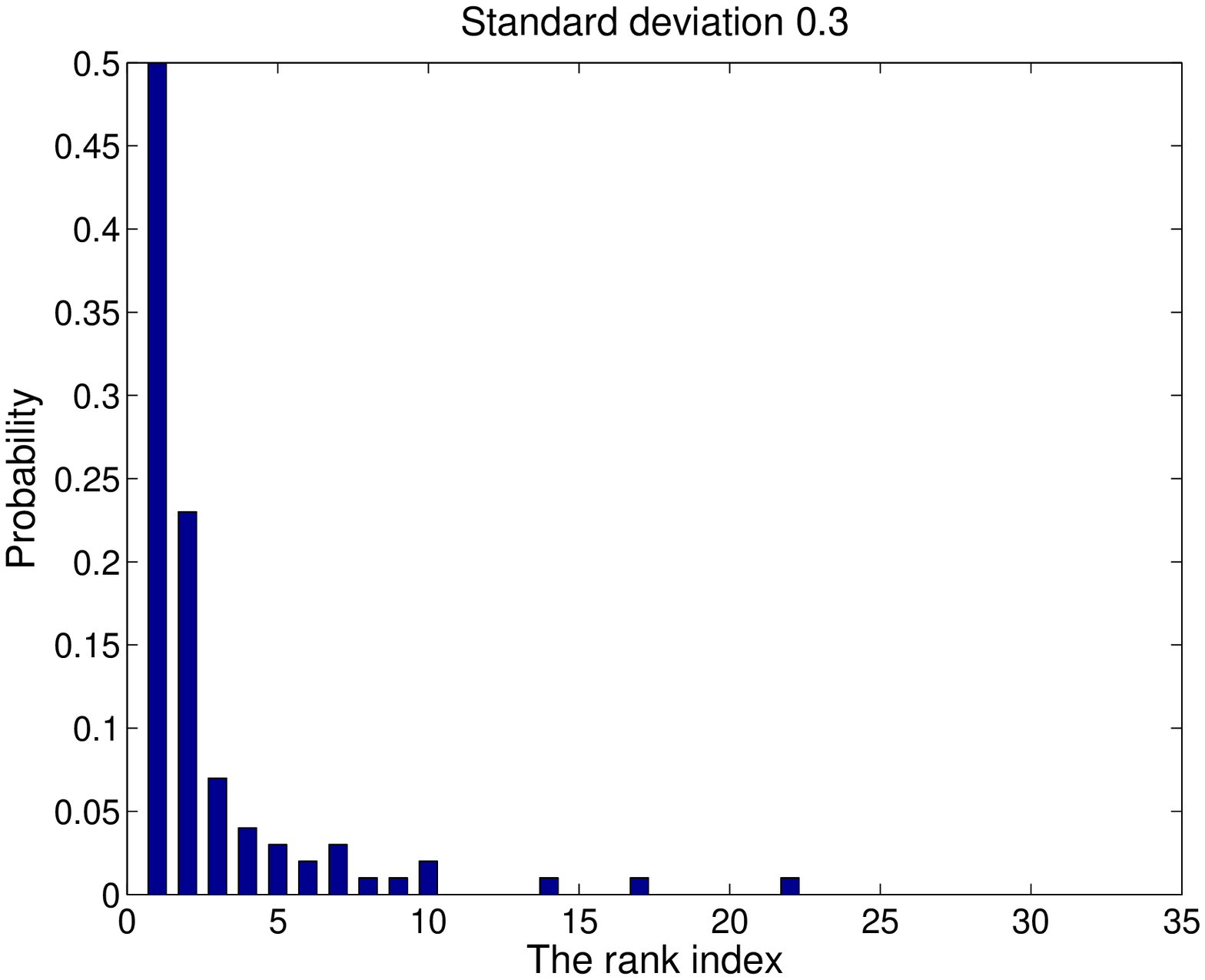}
\caption{The probability distribution of the objective structure's rank with std 0.3.}%
\centering
\label{fig:Rank_std_0dian3}}%
\end{center}
\end{figure}

Figures~\ref{fig:Appearance_ratio_0dian3_onefig} and
\ref{fig:Rank_std_0dian3} show our simulation results for noise with a
standard deviation of $0.3$. The majority of structures still appear
with extremely low probabilities, though the rank of the true structure displays a more scattered distribution because of the increased noise.

%\begin{figure}[!t]
%	\centering
%	\includegraphics[width=0.5\columnwidth]{Fig3}
%	\caption{Average number of structures in each interval with std 0.1}
%	\centering
%	\label{fig:Appearance_ratio_0dian1_onefig}
%\end{figure}
%
%\begin{figure}[!t]
%	\centering
%	\includegraphics[width=0.5\columnwidth]{Fig4}
%	\caption{The probability distribution of the objective structure's rank with std 0.1}
%	\centering
%	\label{fig:Rank_std_0dian1}
%\end{figure}

%\begin{figure}[!t]
%	\centering
%	\includegraphics[width=0.5\columnwidth]{Fig5}
%	\caption{Average number of structures in each interval with std 0.3}	
%	\centering
%	\label{fig:Appearance_ratio_0dian3_onefig}
%\end{figure}
%
%\begin{figure}[!t]
%	\centering
%	\includegraphics[width=0.5\columnwidth]{Fig6}
%	\caption{The probability distribution of the objective structure's rank with std 0.3}
%	\centering
%	\label{fig:Rank_std_0dian3}
%\end{figure}

\begin{figure}[!t]
	\centering
	\includegraphics[width=0.5\columnwidth]{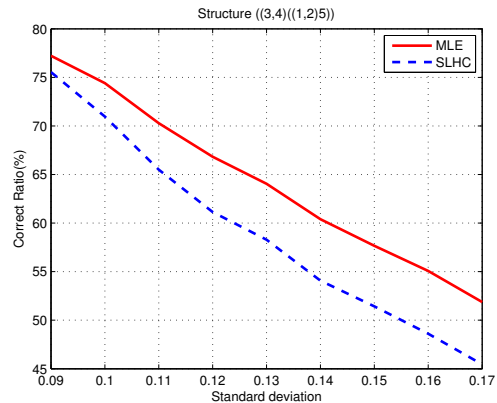}
	\caption{Comparison of MLE and SLHC}
	\centering
	\label{fig:MonteCarloMLEResult}
\end{figure}

To provide an idea of the overall performance of the proposed MLE estimator, Figure~\ref{fig:MonteCarloMLEResult} compares the success rates of the MLE estimator with those of SLHC performed directly on the measurements. In this example, $16$ height vectors were generated uniformly and, for each height vector, a dendrogram $\bm{u}$ with the indicated structure was created. For measurements, $1000$ metrics were sampled from $\slhcinv{\bm{u}}$, and for each of them a measurement $\bm{x}$, drawn according to our data generation model with the prescribed standard deviation. For each $\bm{x}$, we first restricted attention to the most highly ranked $20$ structures. Finally, the  MLE estimator and SLHC were run on $\bm{x}$ and for each the frequency of instances of successful identification of the initial $\bm{\tau}$ were recorded. These ratios were then averaged over the $16$ heights. Figure~\ref{fig:MonteCarloMLEResult} indicates that, on average, the proposed MLE estimator has a better error performance than SLHC, especially so for higher measurement noise variance.

\section{Conclusion}
This paper introduces a rigorous MLE approach to statistical
estimation of SLHC under fairly general assumptions regarding the data
generation process. Simulations with  $5$ particles demonstrate that
the current approach used  in all applications of SLHC~ ---~
calculating SLHC directly from measured data~ ---~ is significantly
outperformed by the  MLE estimation method. 

A clear weakness of our current approach is its computational
complexity which, as presented here, increases very rapidly with data
size. This is largely because of the increased number of MSTs and the
problem of sampling metrics in high dimensions, even though mitigated
by our introduction of MCMC to cull the vast majority of
structures. Further reducing the population of ``MLE-eligible" MSTs is
necessary, and suitable models are being  investigated by us. It is likely that some variant on Kruskal moves, such as~\cite{conf_alenex_OsipovSS09}, will play a useful role here in providing a means to navigate spanning trees more effectively.

We also plan to consider a mixture of ``top-down'' and agglomerative methods of hierarchical approaches to further reduce the complexity of finding the ``top split" in a computationally feasible way, and proceeding recursively from there, thereby reducing the search space in a step-by-step fashion.

\section*{Acknowledgements}
This work was supported by the National Science Fund of China under
Grant 61025006 and 61422114, and by the US Air Force Office of Science
Research under a MURI FA9550-10-1-0567,  and under FA9550-12-1-0418.

\bibliographystyle{gSCS}
\bibliography{SLHC}

\end{document}